\documentclass{amia}
\usepackage[labelfont=bf]{caption}
\usepackage[superscript,nomove]{cite}

\usepackage{color, soul} 
\usepackage{booktabs} 
\usepackage[position=bottom]{subfig} 
\usepackage{siunitx}
\usepackage{slashbox}
\usepackage[ruled]{algorithm2e}
\usepackage{algorithmicx}
\usepackage{graphicx}
\usepackage{textcomp}

\begin{document}

\title{The Impact of Extraneous Variables on the Performance of Recurrent Neural Network Models in Clinical Tasks}

\author{Eugene Laksana, BS, Melissa Aczon, PhD, Long Ho, BS, \\
Cameron Carlin, MS, David Ledbetter, BS, Randall Wetzel, MBBS}
\institutes{
The Laura P. and Leland K. Whittier Virtual Pediatric Intensive Care Unit\\
    Children's Hospital Los Angeles, Los Angeles, California, United States\\
}
\maketitle

\noindent{\bf Abstract}

\textit{Electronic Medical Records (EMR) are a rich source of patient information, including measurements reflecting physiologic signs and administered therapies. Identifying which variables are useful in predicting clinical outcomes can be challenging. Advanced algorithms, such as deep neural networks, were designed to process high-dimensional inputs containing variables in their measured form, thus bypass separate feature selection or engineering steps. We investigated the effect of extraneous input variables on the predictive performance of Recurrent Neural Networks (RNN) by including in the input vector extraneous variables randomly drawn from theoretical and empirical distributions. RNN models using different input vectors (EMR variables only; EMR and extraneous variables; extraneous variables only) were trained to predict three clinical outcomes: in-ICU mortality, 72-hour ICU re-admission, and 30-day ICU-free days. The measured degradations of the RNN's predictive performance with the addition of extraneous variables to EMR variables were negligible.}

\section*{Introduction}
\label{sec:intro}

Electronic Medical Records (EMR) are increasingly adopted by hospitals \cite{henry2016adoption}, resulting in a potential wealth of data for clinical and machine learning research.  A patient's EMR contains comprehensive records of their vital signs, laboratory test results, medications and interventions. Many of these variables may be unrelated to a particular outcome of interest, and in this sense, may be considered \textit{extraneous features} or \textit{noise} for the purposes of modeling that outcome. 

Deep learning (DL) algorithms such as Recurrent Neural Networks (RNNs) were designed to extract salient information from high-dimensional data: the hidden units of each layer are features derived from the input variables to that layer during model training\cite{lecun2015deep}, and this process may be thought of as feature engineering, but automated within the neural network. Combined with regularization techniques such as LASSO regularization, dropout, and recurrent dropout, DL-based models can be robust even when using very high-dimensional input vectors\cite{trevor2009elements, srivastava2014dropout, gal2016theoretically}.  With their feedback loop architecture, RNNs integrate newly acquired data with information retained from previous times to make their decisions.  These characteristics make them attractive and suitable for processing evolving streams of clinical data, as evidenced by their increasing use in medical applications \cite{lipton2015learning, aczon2017dynamic, rajkomar2018scalable, liang2014deep, saqib2018early}. 

Despite the growing popularity of RNNs with clinical data, there is a paucity of literature about the effect of potentially irrelevant data on RNN performance.  We sought to assess this effect by adding extraneous data to RNN model inputs and evaluating the resulting performance.

\section*{Related Works}
\label{sec:related}

Several studies have examined the effect of removing variables and measuring the subsequent effect on model performance. Working on the premise that removing irrelevant or redundant data increases learning accuracy, Khalid, et. al. surveyed several automated feature selection and extraction methods for dimensionality reduction \cite{khalid2014survey}.

Motivated by lowering patient discomfort and financial costs, AlNuaimi, et. al investigated the effect of reducing the number of lab tests on model performance\cite{alnuaimi}. Starting with 35 lab variables as the original inputs for modeling patient deterioration, they repeatedly lowered the number of input variables and employed feature selection algorithms to identify the optimal set of variables. They reported that the Naive Bayes algorithm displayed discrimination improvements when a smaller subset of input features was used. In contrast, the discrimination of the Random Forest, J48 Decision Tree, and Sequential Minimal Optimization models were not improved by the reduction of input data. The study showed that different algorithms respond differently to feature set reduction, and that, excepting Naive Bayes, this reduction neither degraded nor improved their performance.

We previously reported performance comparisons of logistic regression, multilayer perceptron and RNN models for in-ICU mortality using different subsets of EMR variables:  physiologic measurements (vital signs and lab tests) only, therapies only, and all combined\cite{ho2017dependence}. Correlations exist amongst all the variables, and therefore redundancy likely exists when all variables are used. Regardless of algorithm, performance minimally decreased (less than $1\%$ decrease in AUC) when model input was reduced from all variables to physiologic variables only. Across the three algorithms, performance decreased significantly when model input was limited to therapy variables only. These results indicated that the physiologic variables contained most of the relevant information for mortality risk and that the therapy variables contained redundant information, but algorithm performance did not degrade when therapy variables were included in the model.

\section*{Material and Methods}
\label{sec:experiments}

\textit{Clinical Data Sources}

Data were extracted from de-identified observational clinical data collected in Electronic Medical Records (EMR, Cerner) in the Pediatric Intensive Care Unit (PICU) of Children's Hospital Los Angeles (CHLA) between January 2009 and October 2017. A patient record included static information such as demographics, diagnoses, and discharge disposition at the end of an ICU episode. An \textit{episode} is defined as a contiguous admission in the PICU; a patient may have multiple episodes. Each episode also contained irregularly, sparsely and asynchronously \textit{charted} measurements of physiologic observations (e.g. heart rate, blood pressure), laboratory results (e.g. creatine, glucose level), drugs (e.g. epinephrine, furosemide) and interventions (e.g. intubation, oxygen level).  Episodes without discharge disposition were excluded, leaving 7,356 patients with 9,854 episodes.

Prior to any of the computational experiments, the episodes were randomly partitioned into datasets for model training (60\%), validation for hyper-parameter tuning (20\%), and performance evaluation (20\%). To prevent biasing performance evaluation metrics, partitioning was done such that all episodes from a single patient belonged to only one of these sets.

\textit{Target Variables}
\label{sec:target}

We were interested in predicting three clinical outcomes:
\begin{enumerate}
    \item \textbf{Mortality}:  This binary task predicts in-ICU survival or death.  The top portion of Table \ref{tab:targets_and_data} summarizes the number of episodes and mortality rates in the three datasets (training, validation and testing).
    \item \textbf{72-hour ICU Re-admission}: This binary task predicts whether or not a patient was re-admitted to the ICU within 72 hours after physical discharge. Episodes where patients died, were transferred to other ICUs, or were moved to a different hospital were excluded from this experiment. The middle portion of Table \ref{tab:targets_and_data} describes this outcome in the three datasets after the exclusion criteria were applied.
    \item \textbf{30-day ICU-free Days}: This regression task predicts the number of days that a patient was not in the ICU in the 30-day window following a particular time of interest. A patient who died within that window was assigned 0 ICU-free days. Episodes where patients were transferred to the operating room, another ICU, or another hospital were excluded from this experiment. The bottom portion of Table \ref{tab:targets_and_data} describes this outcome in the three datasets after the exclusion criteria were applied. Note that the number of ICU-free days was computed at different time points after ICU admission.
\end{enumerate}

\begin{table*}
\caption{Descriptive statistics of datasets used for each clinical outcome.}
    \begin{center}
    \begin{tabular}{|l|l|l|l|}
    \hline
                    & Train & Valid & Test  \\ \hline
    \textbf{Mortality} &&& \\ 
    \hspace{.3cm} \# Episodes     & 5913  & 1979  & 1962  \\
    \hspace{.3cm} Mortality rate    & 4.19\%  & 3.39\%  & 4.33\%  \\ \hline
    \textbf{72-hour ICU Re-admission} &&& \\ 
    \hspace{.3cm} \# Episodes     & 5556  & 1883  & 1839  \\
    \hspace{.3cm} Re-admission rate  & 2.21\%  & 1.81\%  & 2.28\%  \\ 
    \hspace{.3cm} Median length of stay (hours) & 51.9 & 56.6 & 53.2\\ \hline
    \textbf{30-day ICU-Free Days} &&& \\ 
    \hspace{.3cm} \# Episodes     & 5828  & 1953  & 1934  \\
    \hspace{.3cm} Mean ICU-Free days & & & \\
    \hspace{1.1cm}  @ 3rd hour & 24.91 & 25.10 & 24.71 \\
    \hspace{1.1cm} @ 6th hour      & 25.02 & 25.20 & 24.83 \\
    \hspace{1.1cm} @ 9th hour      & 24.13 & 25.28 & 24.90 \\
    \hspace{1.1cm} @ 12th hour     & 25.20 & 25.33 & 24.94 \\
    \hspace{1.1cm} @ 24th hour     & 25.03 & 25.12 & 24.65 \\ \hline
    \end{tabular}
    \label{tab:experiments_targets}
\end{center}
\label{tab:targets_and_data}
\end{table*}

\textit{Input Variables}

\begin{figure}
    \centering
        \includegraphics[width=3.2in]{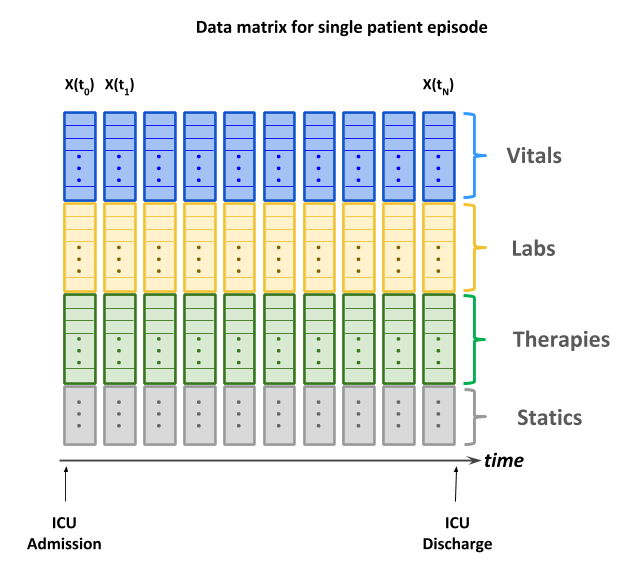}
    \caption{Matrix representation for a single patient episode.  
    A row of data corresponds to measured and imputed values of a single feature, while a column contains values of all features at a single point of time. Static information such as gender is repeated at each time point. Adapted with permission\cite{aczon2017dynamic}.} \vspace{0.6cm}
    \label{fig:data}
\end{figure}

To leverage existing deep learning frameworks, each patient episode's data were first converted to a matrix format illustrated in Figure \ref{fig:data}. Details of this conversion process, which includes z-normalization and forward-fill imputation, are described in previous work \cite{aczon2017dynamic,ho2017dependence}.  A list of all 392 variables in this baseline matrix (whose rows we refer to as \textit{EMR variables} in the remainder of this paper) can be found in the Appendix.
To simulate extraneous, i.e. irrelevant, features as model inputs, rows containing \textit{artificial} data were generated by randomly drawing values from two categories of distributions.

\begin{enumerate}
    \item \textbf{Theoretical}: normal, log-normal, multivariate-normal, exponential, and power function distributions.
    \item \textbf{Empirical}: population-level distributions of each of the 392 variables in the original patient matrix. Feature generation from these distributions is formulized in Algorithm \ref{algo:empirical}.
\end{enumerate}

At each time point in a patient episode matrix, three draws of artificial data were generated from each distribution.  This process generated 1191 rows of \textit{extraneous} features ((5 theoretical x 3) + (392 empirical x 3)) for each patient episode.  Three different types of inputs were then used to train models:  EMR variables only, EMR and extraneous variables, and extraneous variables only. In the second type, the 1191 rows of extraneous variables were appended to the original patient episode matrix, i.e. they comprised 75\% of the input data.

\begin{algorithm}[ht]
\SetAlgoLined
\textit{F} = set of theoretical and empirical feature distributions \\
\textit{E} = set of all episodes, with each episode being a matrix as described in Figure \ref{fig:data} \\

 \For{$f \in F$}{
  \For{$e \in E$}{
    $\|e\| = $ number of recordings in episode $e$\\
    \For{$i \in range(0,3)$}{
        $X$ = random draw of $\|e\|$ samples from $f$ \\
        mask random values in X to emulate missing data frequency in $f$ \\
        add $X$ to $e$'s feature set
    }
  }
}
\caption{Generating extraneous features}
\label{algo:empirical}
\end{algorithm}

\textit{Model Development and Assessment}

RNN models with Long Short-Term Memory (LSTM) architecture \cite{hochreiter1997long} were implemented and trained using the Keras Deep Learning library\cite{keras2018}. The models were trained to make predictions at each time point where a measurement was available.  Model weights were derived from the training set, while hyper-parameters (Table \ref{tab:hyperparam}) were optimized using the validation set. For each clinical outcome, three iterations of RNN model training were performed to better evaluate performance, where each iteration was defined by a different random seed for model weight initialization.

\begin{table}[t]
\caption{Hyper-parameters of RNN models for the three clinical outcomes.}
\begin{tabular}{|l|l|l|l|}
\hline
                            & Mortality            & 72-hour Readmission  & ICU-Free Days      \\ \hline
Number of LSTM Layers       & 3                    & 2                    & 2                  \\
Hidden Units in LSTM Layers & 128, 256, 256        & 420, 375             & 512, 375           \\ 
Batch Size                  & 128                  & 100                  & 100                \\ 
Learning Rate               & 1e-5                 & 1e-4                 & 1e-4               \\ 
Loss                        & binary cross entropy & binary cross entropy & mean squared error \\ 
Optimizer                   & rmsprop              & rmsprop              & rmsprop            \\ 
Dropout                     & 0.2                  & 0.2                  & 0.2                \\ 
Recurrent Dropout           & 0.2                  & 0.2                  & 0.2                \\ 
Regularizer                 & 1e-4                 & 1e-4                 & 1e-6               \\ 
Output Activation           & sigmoid              & sigmoid              & linear             \\ \hline
\end{tabular}
\label{tab:hyperparam}
\end{table}

To facilitate analysis, non-RNN models were also developed. The number of recorded measurements between ICU admission and a particular time of interest was used as an in-ICU mortality risk predictor at that specific time. For the 30-day ICU-free days task, the mean target value derived from the training set (see bottom portion of `train' column in Table 1) was used as a baseline model. 

The mortality models were assessed using area under the receiver operating characteristic curve (AUC) score at 3, 6, 9, 12, and 24 hours following ICU admission.  For the 72-hour readmission task, the models were evaluated via AUC at the time of discharge from the ICU.  The ICU-free days models were assessed by computing the mean absolute error (MAE) of their predictions at 3, 6, 9, 12, and 24 hours following ICU admission.

\section*{Results}
\label{sec:results}

\begin{figure*}
    \centering
    \begin{tabular}{cc}
        \subfloat{\includegraphics[width = 3.0in]{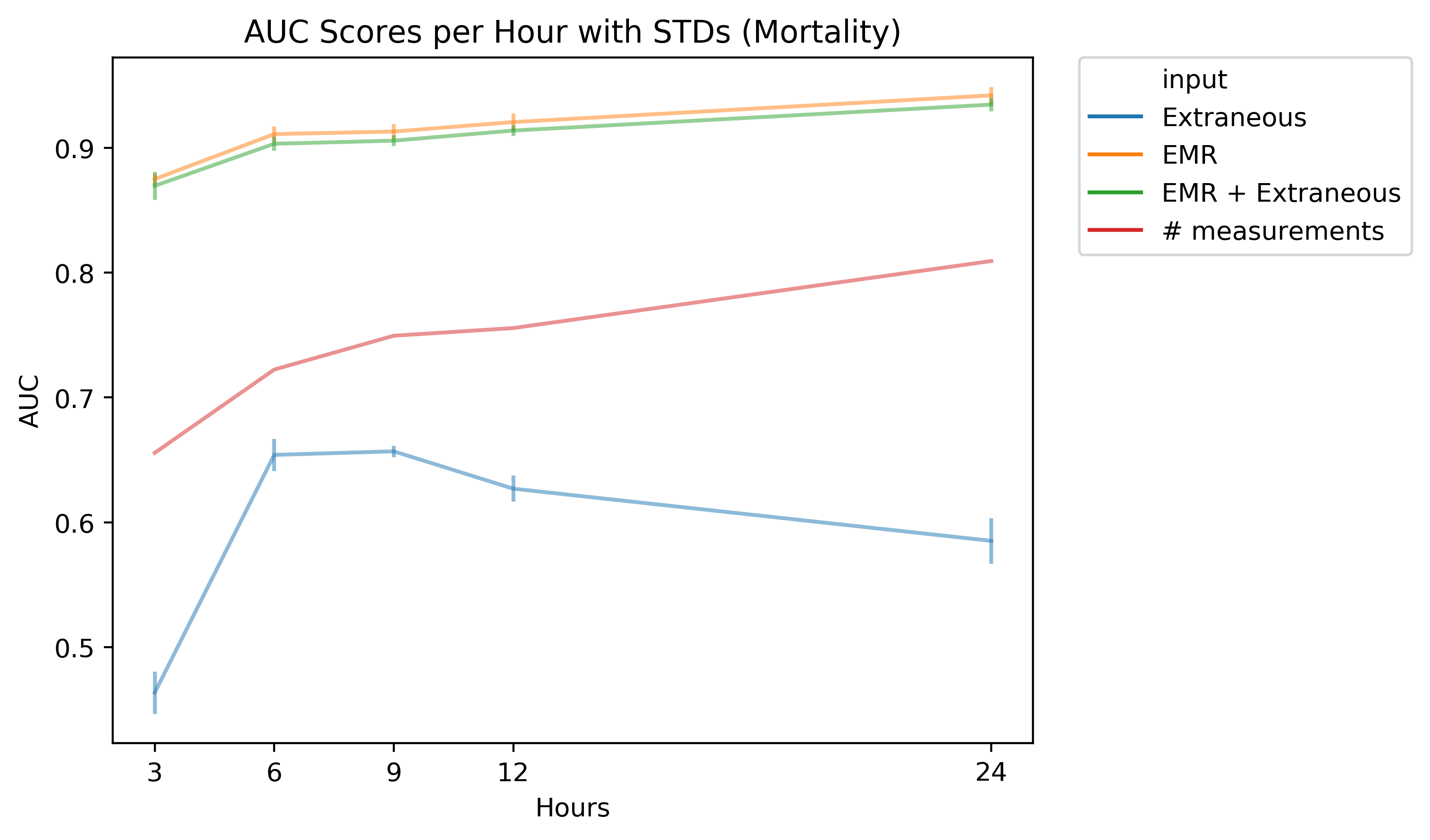}} &
        \subfloat{\includegraphics[width = 3.0in]{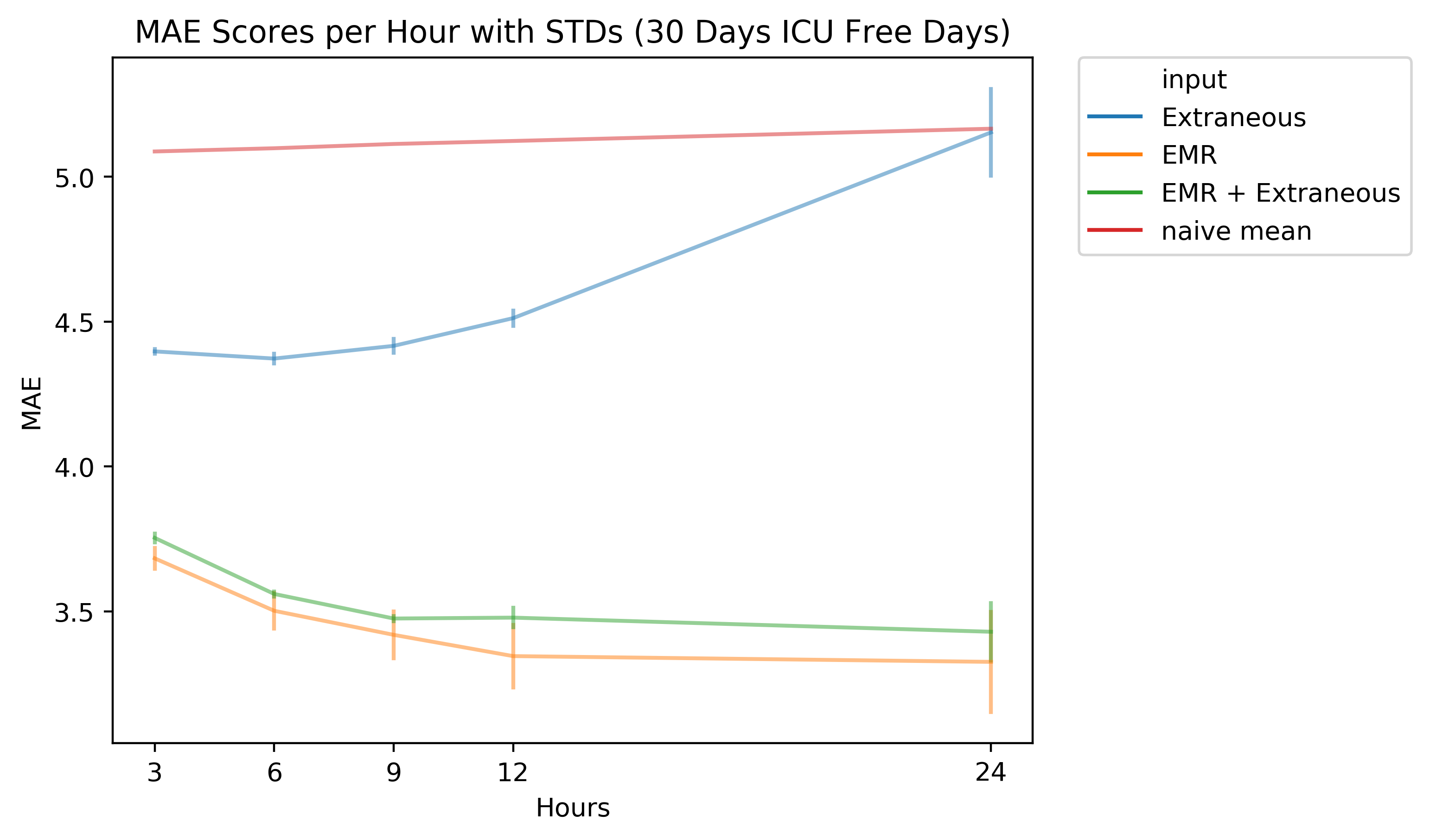}}
    \end{tabular}
\caption{Mortality model AUCs (left) and 30-day ICU-free days models' MAEs (right) with standard deviation lines plotted vs prediction time.}
\label{fig:time_plots}
\end{figure*}

 All performance metrics reported here were computed on the test set. Table \ref{ref:results} displays the mean and standard deviation of model performance from the three training iterations for each outcome. Figure \ref{fig:time_plots} illustrates mortality and ICU-free days model performances as a function of prediction hour. 

\textit{In-ICU Mortality Task.}  The RNN model using only the extraneous variables had the lowest discrimination, with AUCs ranging from 0.46 to 0.66 and peaking between the 6th and 9th hours. The simple model using only the number of measurements between ICU admission and prediction time achieved AUCs ranging from 0.66 to 0.81. The RNN model using only EMR variables had AUCs ranging from 0.870 (at the 3rd hour) to 0.935 (at the 24th hour).  Adding extraneous variables to the EMR variables decreased the RNN's AUCs anywhere from 0.005 to 0.008, representing 0.57\% to 0.89\% degradation in performance. For all input types except for extraneous variables only, longer observation time resulted in higher AUCs.

\textit{72-Hour ICU Re-admission.}  The model using only the extraneous features attained an AUC of 0.489, while the model using only EMR variables had an AUC of 0.644. This AUC did not change when extraneous variables were added to the EMR variables.

\textit{30-Day ICU-Free Days.}  The MAE of the baseline reference value was about 5 days across all prediction times, corresponding to about 20\% of the mean target value. The RNN model using only extraneous features reduced this baseline MAE by more than half a day, except at the 24th hour, where the difference was insignificant. The RNN model using only EMR features had MAE ranging from 3.3 to 3.7 days. Adding extraneous variables to the EMR features increased the MAE by small fractions of a day (0.056 to 0.133, representing 1.6\% to 4\% performance degradation); these two RNN models saw their MAE decrease with longer observation time.

\begin{table*}[htp]
\caption{Model performances on the three target outcomes}
    \begin{center}
    \subfloat[Mortality Model AUC]{
    \begin{tabular}{|l|lllll|}
    \hline
    \backslashbox{Input}{Hour} & 3 & 6 & 9 & 12 & 24 \\
    \hline
    Extraneous only& 0.463 \textpm 0.015 & 0.654 \textpm 0.011 & 0.657 \textpm 0.003 & 0.627 \textpm 0.009 & 0.585 \textpm 0.017 \\
    EMR only & 0.875 \textpm 0.003 & 0.911 \textpm 0.005 & 0.913 \textpm 0.004 & 0.921 \textpm 0.005 & 0.942 \textpm 0.005 \\
    EMR + Extraneous & 0.870 \textpm 0.010 & 0.903 \textpm 0.004 & 0.906 \textpm 0.003 & 0.914 \textpm 0.003 & 0.935 \textpm 0.004 \\
    Recording freq. & 0.656 \textpm 0.000 & 0.722 \textpm 0.000 & 0.750 \textpm 0.000 & 0.756 \textpm 0.000 & 0.809 \textpm 0.000 \\
    \hline
    \end{tabular}
    \label{tab:results_mortality}
    }

    \subfloat[72-Hour Readmission Model AUC]{
        \begin{tabular}{|l|l|}
            \hline
            Input & AUC\textpm std \\ \hline
            Extraneous only & 0.489 \textpm  0.008 \\
            EMR only & 0.644 \textpm 0.016 \\ 
            EMR + Extraneous & 0.644 \textpm 0.017 \\ \hline
        \end{tabular}
    \label{tab:results_read}
    } 
    
    \subfloat[30-Day ICU-Free Days Model MAE (in days)]{
        \begin{tabular}{|l|lllll|}
        \hline
        \backslashbox{Input}{Hour} & 3 & 6 & 9 & 12 & 24 \\
        \hline
        Extraneous only & 4.397 \textpm 0.009 & 4.373 \textpm 0.016 & 4.416 \textpm 0.024 & 4.512 \textpm 0.026 & 5.153 \textpm 0.150 \\
        EMR only & 3.683 \textpm 0.036 & 3.502 \textpm 0.061 & 3.419 \textpm 0.081 & 3.345 \textpm 0.108 & 3.326 \textpm 0.173 \\
        EMR + Extraneous & 3.754 \textpm 0.015 & 3.560 \textpm 0.009 & 3.475 \textpm 0.009 & 3.478 \textpm 0.033 & 3.429 \textpm 0.100 \\
        Target mean from train set & 5.088 \textpm 0.000 & 5.099 \textpm 0.000 & 5.113 \textpm 0.000 & 5.124 \textpm 0.000 & 5.166 \textpm 0.000 \\
        \hline
    \end{tabular}
    \label{tab:results_free}
    }

\end{center}
\label{ref:results}
\ref{sec:target}
\end{table*}

\section*{Discussion}
\label{sec:discussion}

Including extraneous variables alongside true EMR variables, with the extraneous features comprising 75\% of the input, degraded the RNN's performance only slightly:  less than 1\% on the in-ICU mortality and re-admission tasks, and 2\%-4\% on the 30-day ICU-free days task.  Even on the third task, the difference meant small fractions in absolute days. These results demonstrate the RNN's ability to manage extraneous information when predicting the clinical outcomes. Incorporating techniques such as dropout, recurrent dropout, and LASSO regularization help mitigate overfitting effects when high dimensional data are involved \cite{trevor2009elements,srivastava2014dropout}. Recent work has even suggested that, when done properly, adding noise to models can be a regularization technique \cite{noh2017regularizing}.

Not surprisingly, among the RNN models, those using only extraneous variables as inputs had the worst performance across all outcomes. The RNN model using only extraneous variables to predict ICU re-admission displayed random discrimination. The RNN model using only extraneous variables to predict ICU-free days performed about the same or slightly better than the baseline reference, indicating that this RNN model learned the population mean. For the in-ICU mortality task, the RNN model using only extraneous variables performed better than random (AUC $>0.5$). Previous research has shown that nurse charting frequencies reflect clinicians' anticipation of clinical outcome \cite{donabedian1966evaluating, donabedian1988quality}. These findings are consistent with the range of AUCs (0.65 to 0.81) from a classifier that used only the number of recorded measurements between ICU admission and prediction time (fourth row of Table \ref{tab:results_mortality}). By construction, the temporal sampling of the extraneous features matched the temporal sampling of the charted measurements, which may explain the better than random performance of the RNN mortality model that used only extraneous features. This would suggest that the RNN learned some correlation between charting frequency and mortality risk from random values that had nothing to do with an individual patient but, when presented as a sequence, implicitly contained nurse charting frequency for that patient.

This study is limited by the single-center nature of the data used in the experiments. Future work will extend these experiments to other clinical tasks and algorithms, such as the multilayer perceptron, random forest, and logistic regression, to assess these algorithms' robustness against extraneous or superfluous data.

\section*{Conclusion}
\label{sec:conclusion}

This study demonstrated that RNN models with LSTM architecture can robustly manage high-dimensional data even when the majority of that data contain irrelevant information. The experiments focused on three clinical outcomes:  in-ICU mortality, 72-hour ICU re-admission, and ICU-free days. RNNs can be trained for these clinically relevant tasks without model developers spending additional meticulous efforts on feature selection.

\section*{Acknowledgements}
The authors are grateful to the L. K. Whittier Foundation for supporting this work.


\makeatletter
\renewcommand{\@biblabel}[1]{\hfill #1.}
\makeatother

\bibliographystyle{vancouver}
\bibliography{Includes/rnn_noise.bib}

\vspace{0.5cm}
\pagebreak
\section*{Appendix}
\vspace{0.5cm}
\begin{table}[htp]
\caption{EMR variables (demographics, vitals and labs) in patient episode matrix. Demographics such as gender and race/ethnicity were encoded as binary variables.}
\centering
\resizebox{\textwidth}{!}{
\begin{tabular}{lllll}\hline
\multicolumn{4}{c} {Demographics and Vitals}\\ \hline
Age & Sex\_F & Sex\_M & race\_African American \\
race\_Asian/Indian/Pacific Islander  & race\_Caucasian/European Non-Hispanic & race\_Hispanic & race\_unknown \\ 
Abdominal Girth & FLACC Pain Face & Left Pupillary Response Level & Respiratory Effort Level \\
Activity Level & FLACC Pain Intensity & Level of Consciousness & Respiratory Rate \\
Bladder pressure & FLACC Pain Legs & Lip Moisture Level & Right Pupil Size After Light \\
Capillary Refill Rate & Foley Catheter Volume & Mean Arterial Pressure & Right Pupil Size Before Light \\
Central Venous Pressure & Gastrostomy Tube Volume & Motor Response Level & Right Pupillary Response Level \\
Cerebral Perfusion Pressure & Glascow Coma Score & Nasal Flaring Level & Sedation Scale Level \\
Diastolic Blood Pressure & Head Circumference & Near-Infrared Spectroscopy SO2 & Skin Turgor\_edema \\
EtCO2 & Heart Rate & Nutrition Level & Skin Turgor\_turgor \\
Extremity Temperature Level & Height & Oxygenation Index & Systolic Blood Pressure \\
Eye Response Level & Hemofiltration Fluid Output & PaO2 to FiO2 & Temperature \\
FLACC Pain Activity & Intracranial Pressure & Patient Mood Level & Verbal Response Level \\
FLACC Pain Consolability & Left Pupil Size After Light & Pulse Oximetry & WAT1 Total \\
FLACC Pain Cry & Left Pupil Size Before Light & Quality of Pain Level & Weight \\ \\ \hline
\multicolumn{4}{c} {Labs}\\ \hline
ABG Base excess & CBG PCO2 & GGT & Neutrophils \% \\
ABG FiO2 & CBG PO2 & Glucose & PT \\
ABG HCO3 & CBG TCO2 & Haptoglobin & PTT \\
ABG O2 sat & CBG pH & Hematocrit & Phosphorus level \\
ABG PCO2 & CSF Bands\% & Hemoglobin & Platelet Count \\
ABG PO2 & CSF Glucose & INR & Potassium \\
ABG TCO2 & CSF Lymphs \% & Influenza Lab & Protein Total \\
ABG pH & CSF Protein & Lactate & RBC Blood \\
ALT & CSF RBC & Lactate Dehydrogenase Blood & RDW \\
AST & CSF Segs \% & Lactic Acid Blood & Reticulocyte Count \\
Albumin Level & CSF WBC & Lipase & Schistocytes \\
Alkaline phosphatase & Calcium Ionized & Lymphocyte \% & Sodium \\
Amylase & Calcium Total & MCH & Spherocytes \\
Anti-Xa Heparin & Chloride & MCHC & T4 Free \\
B-type Natriuretic Peptide & Complement C3 Serum & MCV & TSH \\
BUN & Complement C4 Serum & MVBG Base Excess & Triglycerides \\
Bands \% & Creatinine & MVBG FiO2 & VBG Base excess \\
Basophils \% & Culture Blood & MVBG HCO3 & VBG FiO2 \\
Bicarbonate Serum & Culture CSF & MVBG O2 Sat & VBG HCO3 \\
Bilirubin Conjugated & Culture Fungus Blood & MVBG PCO2 & VBG O2 sat \\
Bilirubin Total & Culture Respiratory & MVBG PO2 & VBG PCO2 \\
Bilirubin Unconjugated & Culture Urine & MVBG TCO2 & VBG PO2 \\
Blasts \% & Culture Wound & MVBG pH & VBG TCO2 \\
C-Reactive Protein & D-dimer & Macrocytes & VBG pH \\
CBG Base excess & ESR & Magnesium Level & White Blood Cell Count \\
CBG FiO2 & Eosinophils \% & Metamyelocytes \% & \\
CBG HCO3 & Ferritin Level & Monocytes \% &  \\
CBG O2 sat & Fibrinogen & Myelocytes \% &  \\ \hline
\end{tabular}
}

\label{rnn-inputs_5col}
\end{table}


\begin{table}
\caption{EMR Variables (drugs and interventions) in baseline patient episode matrix.}
\centering
\resizebox{\textwidth}{!}{
\begin{tabular}{lllll}\hline
\multicolumn{4}{c} {Drugs}\\ \hline
Acetaminophen/Codeine\_inter & Clonazepam\_inter & Ipratropium Bromide\_inter & Oseltamivir\_inter \\
Acetaminophen/Hydrocodone\_inter & Clonidine HCl\_inter & Isoniazid\_inter & Oxacillin\_inter \\
Acetaminophen\_inter & Cyclophosphamide\_inter & Isradipine\_inter & Oxcarbazepine\_inter \\
Acetazolamide\_inter & Desmopressin\_inter & Ketamine\_cont & Oxycodone\_inter \\
Acyclovir\_inter & Dexamethasone\_inter & Ketamine\_inter & Pantoprazole\_inter \\
Albumin\_inter & Dexmedetomidine\_cont & Ketorolac\_inter & Penicillin G Sodium\_inter
\\
Albuterol\_inter & Diazepam\_inter & Labetalol\_inter & Pentobarbital\_inter \\
Allopurinol\_inter & Digoxin\_inter & Lactobacillus\_inter & Phenobarbital\_inter \\
Alteplase\_inter & Diphenhydramine HCl\_inter & Lansoprazole\_inter & Phenytoin\_inter
\\
Amikacin\_inter & Dobutamine\_cont & Levalbuterol\_inter & Piperacillin/Tazobactam\_inter \\
Aminophylline\_cont & Dopamine\_cont & Levetiracetam\_inter & Potassium Chloride\_inter
 \\
Aminophylline\_inter & Dornase Alfa\_inter & Levocarnitine\_inter & Potassium Phosphat
e\_inter \\
Amlodipine\_inter & Enalapril\_inter & Levofloxacin\_inter & Prednisolone\_inter \\
Amoxicillin/clavulanic acid\_inter & Enoxaparin\_inter & Levothyroxine Sodium\_inter &
 Prednisone\_inter \\
Amoxicillin\_inter & Epinephrine\_cont & Lidocaine\_inter & Propofol\_cont \\
Amphotericin B Lipid Complex\_inter & Epinephrine\_inter & Linezolid\_inter & Propofol
\_inter \\
Ampicillin/Sulbactam\_inter & Epoetin\_inter & Lisinopril\_inter & Propranolol HCl\_inter \\
Ampicillin\_inter & Erythromycin\_inter & Lorazepam\_inter & Racemic Epi\_inter \\
Aspirin\_inter & Factor VII\_inter & Magnesium Sulfate\_inter & Ranitidine\_inter \\
Atropine\_inter & Famotidine\_inter & Meropenem\_inter & Rifampin\_inter \\
Azathioprine\_inter & Fentanyl\_cont & Methadone\_inter & Risperidone\_inter \\
Azithromycin\_inter & Fentanyl\_inter & Methylprednisolone\_inter & Rocuronium\_inter \\
Baclofen\_inter & Ferrous Sulfate\_inter & Metoclopramide\_inter & Sildenafil\_inter \\
Basiliximab\_inter & Filgrastim\_inter & Metronidazole\_inter & Sodium Bicarbonate\_inter \\
Budesonide\_inter & Fluconazole\_inter & Micafungin\_inter & Sodium Chloride\_inter \\
Bumetanide\_inter & Fluticasone\_inter & Midazolam HCl\_cont & Sodium Phosphate\_inter \\
Calcium Chloride\_cont & Fosphenytoin\_inter & Midazolam HCl\_inter & Spironolactone\_inter \\
Calcium Chloride\_inter & Furosemide\_cont & Milrinone\_cont & Sucralfate\_inter \\
Calcium Gluconate\_inter & Furosemide\_inter & Montelukast Sodium\_inter & Tacrolimus\_inter \\
Carbamazepine\_inter & Gabapentin\_inter & Morphine\_cont & Terbutaline\_cont \\
Cefazolin\_inter & Ganciclovir Sodium\_inter & Morphine\_inter & Tobramycin\_inter \\
Cefepime\_inter & Gentamicin\_inter & Mycophenolate Mofetl\_inter & Topiramate\_inter \\
Cefotaxime\_inter & Glycopyrrolate\_inter & Naloxone HCL\_cont & Trimethoprim/Sulfamethoxazole\_inter \\
Cefoxitin\_inter & Heparin\_cont & Naloxone HCL\_inter & Ursodiol\_inter \\
Ceftazidime\_inter & Heparin\_inter & Nifedipine\_inter & Valganciclovir\_inter \\
Ceftriaxone\_inter & Hydrocortisone\_inter & Nitrofurantoin\_inter & Valproic Acid\_inter \\
Cephalexin\_inter & Hydromorphone\_cont & Nitroprusside\_cont & Vancomycin\_inter \\
Chloral Hydrate\_inter & Hydromorphone\_inter & Norepinephrine\_cont & Vasopressin\_cont \\
Chlorothiazide\_inter & Ibuprofen\_inter & Nystatin\_inter & Vecuronium\_inter \\
Ciprofloxacin HCL\_inter & Immune Globulin\_inter & Octreotide Acetate\_cont & Vitamin K\_inter \\
Cisatracurium\_cont & Insulin\_cont & Olanzapine\_inter & Voriconazole\_inter \\
Clindamycin\_inter & Insulin\_inter & Ondansetron\_inter & \\ \\ \hline
\multicolumn{4}{c} {Interventions}\\ \hline
Abdominal X Ray & Diversional Activity\_tv & NIV Mode & Range of Motion Assistance Type \\
Arterial Line Site & ECMO Hours & NIV Set Rate & Sedation Intervention Level \\
CT Abdomen Pelvis & EPAP & Nitric Oxide & Sedation Response Level \\
CT Brain & FiO2 & Nurse Activity Level Completed & Tidal Volume Delivered \\
CT Chest & Gastrostomy Tube Location & O2 Flow Rate & Tidal Volume Expiratory \\
Central Venous Line Site & HFOV Amplitude & Oxygen Mode Level & Tidal Volume Inspiratory \\
Chest Tube Site & HFOV Frequency & Oxygen Therapy & Tidal Volume Set \\
Chest X Ray & Hemofiltration Therapy Mode & PEEP & Tracheostomy Tube Size \\
Comfort Response Level & IPAP & Peak Inspiratory Pressure & Ventilator Rate \\
Continuous EEG Present & Inspiratory Time & Peritoneal Dialysis Type & Ventriculostomy Site \\
Diversional Activity\_books & MRI Brain & Pharmacological Comfort Measures Given & Visitor Mood Level \\
Diversional Activity\_music & Mean Airway Pressure & Position Support Given & Visitor Present \\
Diversional Activity\_play & Mechanical Ventilation Mode & Position Tolerance Level & Volume Tidal \\
Diversional Activity\_toys & MultiDisciplinaryTeam Present & Pressure Support & 
 \\ \hline
\end{tabular}
}
\label{rnn-inputs_5col_b}
\end{table}

\end{document}